\documentclass[conference]{IEEEtran}
\usepackage{cite}
\usepackage{amsmath,amssymb,amsfonts}
\usepackage{algorithmic}
\usepackage{graphicx}
\usepackage{textcomp}
\usepackage{xcolor}
\usepackage[hyphens]{url}
\usepackage{hyperref}
\hypersetup{
  colorlinks=true,
  linkcolor=black,
  citecolor=black,
  urlcolor=blue,
  pdftitle={LLM-Based Multi-Agent System for Simulating and Analyzing Marketing and Consumer Behavior},
  pdfauthor={Man-Lin (Carol) Chu et al.}
}

\def\BibTeX{{\rm B\kern-.05em{\sc i\kern-.025em b}\kern-.08em
    T\kern-.1667em\lower.7ex\hbox{E}\kern-.125emX}}
\begin{document}

\title{LLM-Based Multi-Agent System for Simulating and Analyzing Marketing and Consumer Behavior}

\author{
  \IEEEauthorblockN{Man-Lin Chu}
  \IEEEauthorblockA{
    School of Business\\
    Clark University\\
    Worcester, MA, USA\\
    mchu@clarku.edu
  }
  \and
  \IEEEauthorblockN{Lucian Terhorst, Kadin Reed, Tom Ni, Weiwei Chen}
  \IEEEauthorblockA{
    Department of Computer Science\\
    Clark University\\
    Worcester, MA, USA\\
    \{lterhorst, kadreed, tni\}@clarku.edu, wei77455@gmail.com
  }
  \and
  \IEEEauthorblockN{Rongyu Lin$^{*}$}
  \IEEEauthorblockA{
    School of Computing and Engineering\\
    Quinnipiac University\\
    Hamden, CT, USA\\
    rongyu.lin@quinnipiac.edu
  }
}

\maketitle
\begin{center}
\small
\textbf{Preprint Notice}\\[3pt]
This is the author-accepted manuscript (AAM) of the paper 
\emph{``LLM-Based Multi-Agent System for Simulating and Analyzing Marketing and Consumer Behavior,''}
accepted for publication in the 
\textbf{IEEE International Conference on e-Business Engineering (ICEBE 2025)}, to be held 10–12 November 2025 at Mustaqbal University, Buraydah, Saudi Arabia.\\[6pt]
© 2025 IEEE. Personal use of this material is permitted. Permission from IEEE must be obtained for all other uses,
including reprinting or republishing, or for creating derivative 
\end{center}
\vspace{0.5em}

\begin{abstract}  
Simulating consumer decision-making is vital for designing and evaluating marketing strategies before costly real-world deployment. However, post-event analyses and rule-based agent-based models (ABMs) struggle to capture the complexity of human behavior and social interaction. We introduce an LLM-powered multi-agent simulation framework that models consumer decisions and social dynamics. Building on recent advances in large language model simulation in a sandbox environment, our framework enables generative agents to interact, express internal reasoning, form habits, and make purchasing decisions without predefined rules. In a price-discount marketing scenario, the system delivers actionable strategy-testing outcomes and reveals emergent social patterns beyond the reach of conventional methods. This approach offers marketers a scalable, low-risk tool for pre-implementation testing, reducing reliance on time-intensive post-event evaluations and lowering the risk of underperforming campaigns.

\end{abstract}

\section{Introduction}
Marketing decision-making has always carried uncertainty. In today's e-business era, dominated by digital advertising, social media trends, real-time personalization, and dynamic pricing, launching untested strategies carries higher stakes than ever. Businesses have traditionally relied on sales data analysis, surveys, focus groups, and binary testing to assess promotional effectiveness, pricing schemes, or product bundles. While these methods generate valuable insights, they require  significant time and resource, and usually provide results after campaign execution, leaving businesses to bear the costs and risks of underperforming strategies.

To address these limitations, researchers in marketing and behavioral economics have tried to simulate business scenarios through agent-based models (ABMs) \cite{macal2006tutorial, rand2011agent}. However, conventional ABMs rely primarily on programmed rules, whereas LLM-powered agents, trained on massive datasets, can learn complex behaviors and generate context-sensitive outputs. They are able to reason, plan, remember, and understand language, making their actions more adaptable and human-like than those of rule-based models. This makes generative agents especially useful for modeling social dynamics such as habit formation, interaction, and word-of-mouth diffusion, key drivers in today's marketing environments, since consumer behavior is rarely linear or isolated; customer journey frameworks have evolved over a century to reflect increasing complexity in consumer decision-making, as illustrated by the progression from AIDA \cite{lewis1898aida} to AIDMA \cite{hall1920aidma}, AISAS \cite{dentsu2004aisas}, and AIDEES \cite{wijaya2012aisdalslove}. These developments highlight the need for simulation approaches that capture the full process of shaping consumer behavior, not just the final purchase decision.

Recent advances in Park et al. proposed the Generative Agents framework, where LLM-powered agents, equipped with memory, planning, reflection, and language interaction, to simulate believable human behavior in a sandbox world \cite{park2023generative}. These agents demonstrate emergent behavior like forming routines, organizing events, and spreading information, making them ideal for simulating marketing phenomena including social influence, habitual choice, and value perception.

Building on this framework, our study explores using generative agents to simulate consumer response to localized marketing promotions. We construct a week-long virtual town experiment where agents plan daily schedules, manage resources, shop with earned income, converse and make social commitments, coordinate visits to town locations, and choose between a cafe, fast-food outlet, and family restaurant for meals. A fried chicken shop offers a midweek 20\% discount while others maintain regular pricing. Like real consumers, agents are influenced by promotional awareness and both external and internal factors including resources (money, groceries, energy, time), and past experiences, as detailed in Section IV.

\begin{figure*}[t]
    \centering
    \includegraphics[width=0.9\linewidth]{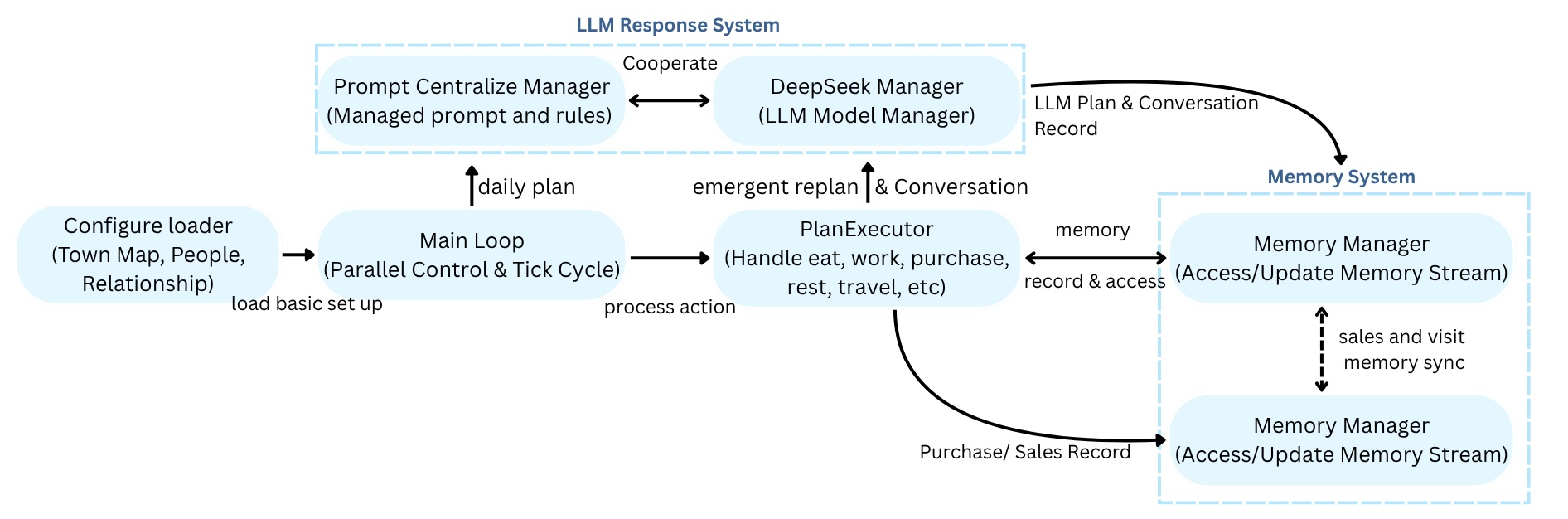}
    \caption{Experiment flow showing how multiple management systems cooperate.}
    \label{fig:flow}
\end{figure*}

We selected price discount strategy as our focus for theoretical and practical reasons. Theoretically, consumer psychology research shows individual variation in deal promotion proneness (DPP) - the tendency of some consumers to respond more strongly to price-based incentives than others \cite{lichtenstein1997deal}. This well-documented effect provides a behavioral baseline for observing agent variation. Practically, discount strategies are easily controlled within complex interactive environments where agents form routines, make trade-offs, and influence each other through conversation based on identity and social class. This makes discount promotions a tractable starting point for constructing virtual consumer ecosystems with emergent dynamics. 

This work investigates whether LLM-powered generative agents can realistically simulate real-world consumer behavior by embedding price strategies in agent environments, providing actionable insights for marketing strategy evaluation through controlled promotional simulations.

\section{Related Work}
Creating believable proxies of human behavior has emerged as a central challenge in human-AI interaction, multi-agent simulation, and computational social science. Despite long-standing efforts, human cognition's inherent complexity and variability make this challenging. Recent LLM advances, combined with memory, planning, and social reasoning architectures, enable interactive human-like agents. Our study connects these innovations with marketing science, grounding consumer behavior simulation in established theories from consumer psychology and behavioral economics.

\subsection{Agent-Based and Generative Modeling}
ABMs have simulated consumer behavior across promotional strategies by deriving macro trends from micro rules\cite{macal2006tutorial, rand2011agent}, yet their dependence on fixed heuristics limits cognitive depth, social influence modeling, and individual variation capture in emotional, motivational, and contextual dimensions. 

Recent advancements in LLMs have enabled the development of generative agents autonomously reason and plan in natural language. Park et al. \cite{park2023generative} introduced the Generative Agents framework, demonstrating how LLM-powered agents equipped with memory, reflection, and planning can generate believable human behavior in a sandbox town. These agents exhibited emergent dynamics such as routine formation, rumor spreading, and collaborative planning. However, this work centers on open-ended scenarios, not domain-specific simulations (e.g., consumer responses to marketing interventions).

\subsection{Consumer Psychology and Promotion Sensitivity}
From a marketing psychology perspective, purchase behavior is not uniform; consumer preferences and promotional responsiveness vary between individuals. Lichtenstein et al. \cite{lichtenstein1997deal} introduced the concept of DPP to explain differential price sensitivity. Although DPP has traditionally been explored through field experiments and survey instruments, few studies have examined whether such behavioral variability can be replicated within generative simulations. Our study bridges this gap by embedding agents with heterogeneous traits and dynamic "persona" (can be seen as profile of a consumer), evaluating their discount responses in interactive environments.

\section{Simulation Setting For Discount Scenario}
Our simulation features 11 agents and 10 locations for residence, dining, shopping, work, and leisure. Each agent plans a daily routine using LLM responses guided by structured prompts and executes it within the simulation, focusing on food-purchase behavior to evaluate discount strategy. Through repeated simulation cycles, we observe how price incentives influence public opinion, consumer stickiness, choice shifts, and purchase decision reasoning.
\begin{figure*}[t]
    \centering
    \includegraphics[width=0.9\linewidth,trim=0cm 0cm 0cm 1.7cm, clip]{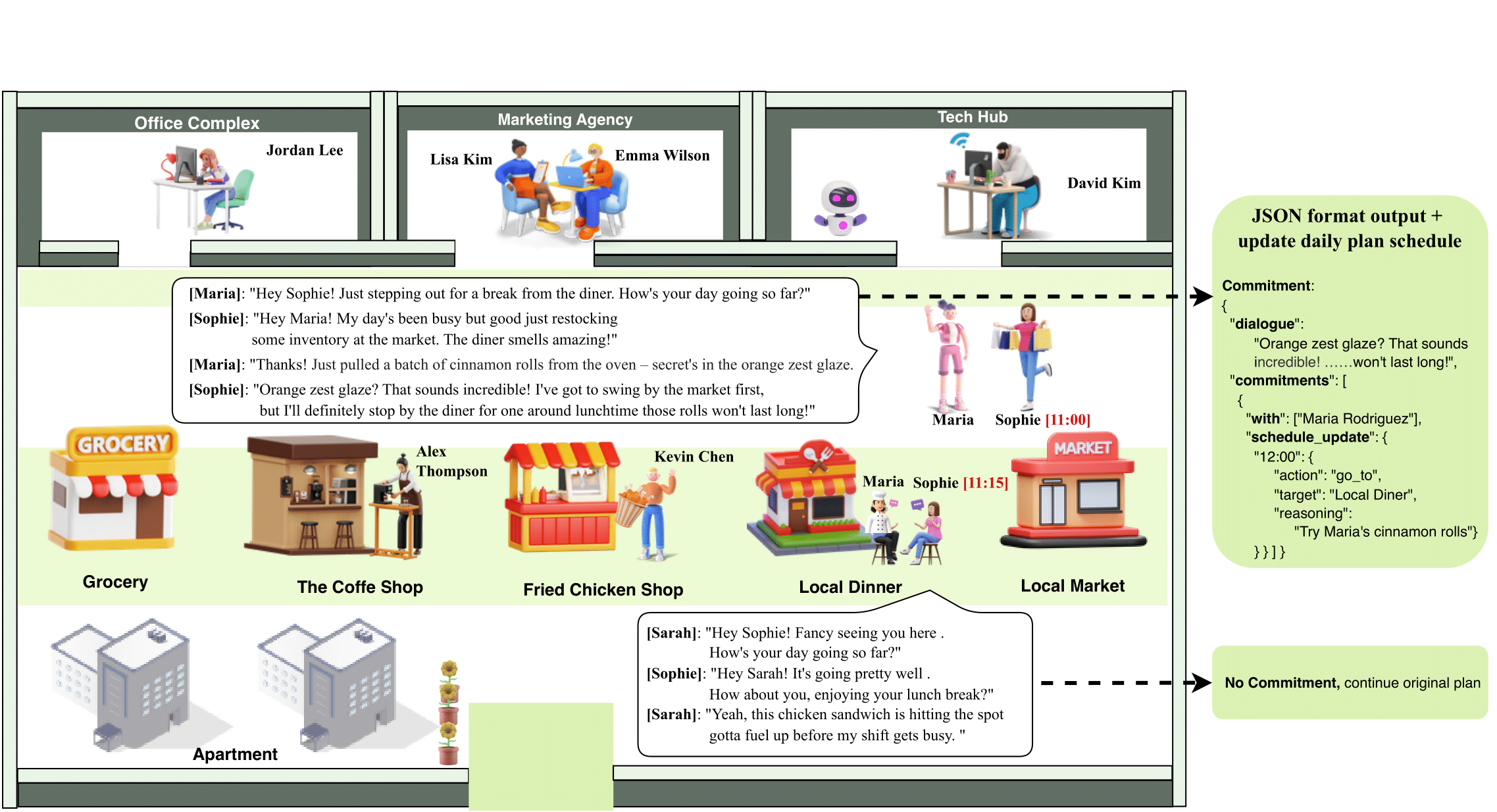}
    \caption{Multi-Agent Simulation In The Town}
    \label{fig:Town}
\end{figure*}
\subsection{LLM-Powered Agent Cognition}
DeepSeek-V3 powers the simulation, enabling agents to plan, execute, and communicate naturally. As Fig.~\ref{fig:flow}, we centralized model running through closed-cycle with execution, memory, and LLM response systems. Agents receive structured prompts retrieving their current state, needs, environment, and memory from memory system. These diverse prompts covered daily planning, dining decisions, interpersonal conversations, and work routines types for easily catching agent needs. DeepSeek model returns structured JSON encoding each agent's action and reasoning, and pass to execution system. For instance:
\vspace{1em}
\begin{quote}
\footnotesize
\ttfamily
\raggedright
"time": 12,\\
"action": "eat",\\
"target": "Local Diner",\\
"description": "Lunch break at nearby diner",\\
"energy\_considerations": "+40 energy from restaurant meal"\\
\end{quote}
\vspace{1em}

\subsection{Agent Architecture and Environment Configuration}
The simulation is initialized using a centralized configuration schema that defines both the agents and the environment. The example picture Fig. \ref{fig:Town} shows the town map layout, physical distance between locations, residential assignments, shop offerings, and agent personas. The design emphasizes real-world constraints and behavioral triggers.

\subsubsection{Town Map and Spatial Dynamics}
The town is spatially segmented into three primary areas: Residential, Business District, and Main Street, a total of 10 locations available for agents to freely visit through prompts. Shops such as grocery shop, Fried Chicken Shop (dining), and The Coffee Shop (dining) are deliberately placed along a central path to maximize casual exposure. Our experiment focus, the only shop implements discount event, Fried Chicken Shop, is at [0.0] grid, the center of the whole town map. Agents navigate predefined paths between work, home and leisure, ensuring passage through commercial areas and purchase consideration.

\vspace{1em}
\begin{quote}
\footnotesize
\ttfamily
\raggedright
"Fried Chicken Shop": [0, 0],\\
"The Coffee Shop": [-3, 0],\\
"Local Diner": [3, 0],\\
"travel\_paths": [[-3, 0], [0, 0], [3, 0]]\\
\end{quote}
\vspace{1em}

\subsubsection{Persona and Socioeconomic Diversity}
\begin{figure*} [t]
    \centering
    \includegraphics[width=0.65\linewidth,trim=0cm 0cm 0cm 2.5cm]{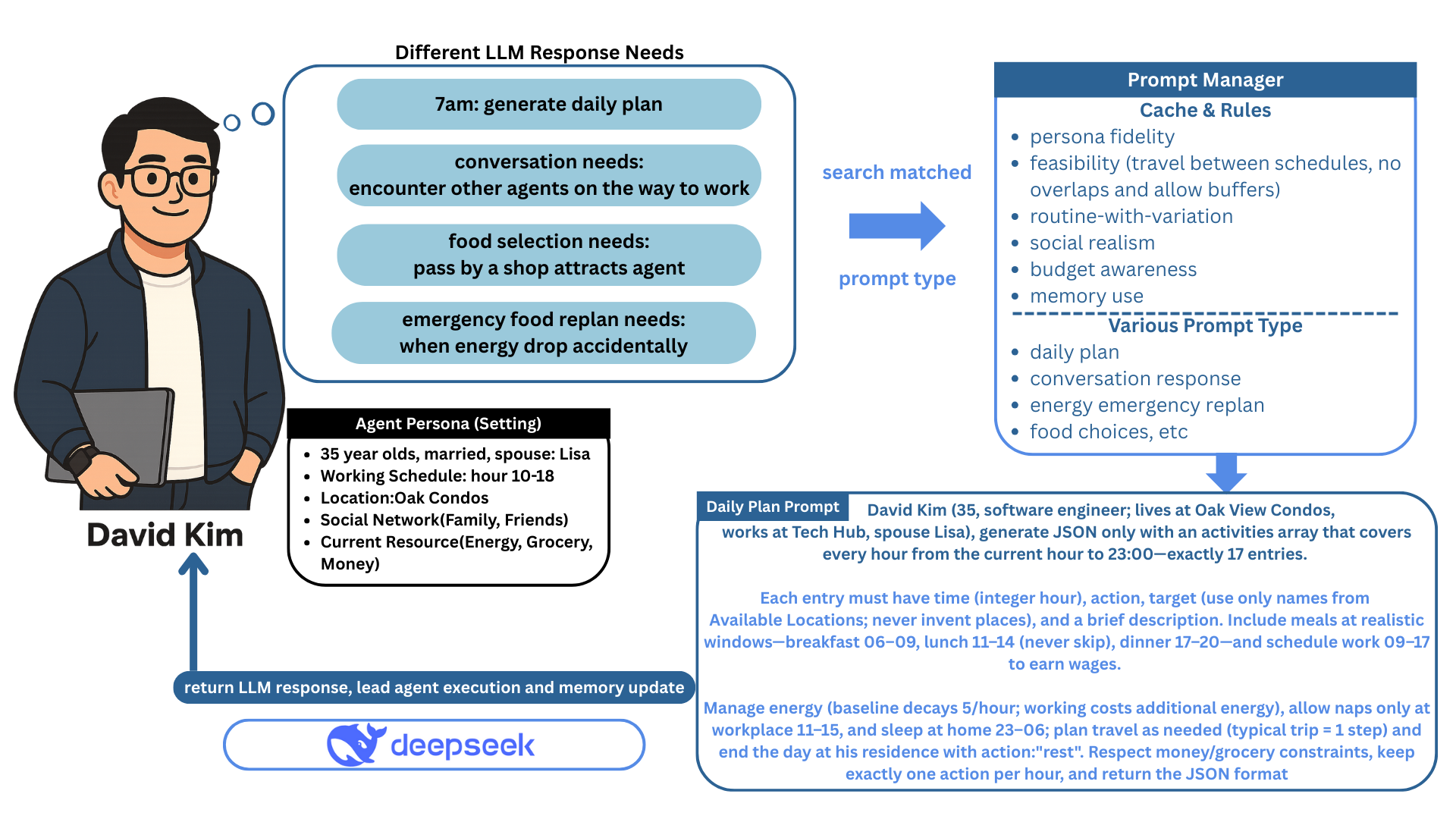}
    \caption{Prompt Engineer Process within Prompt Manager and Deepseek Model: from Needs, Trigger, to Response}
    \label{fig:Prompt}
\end{figure*}

Eleven agents cover diverse demographics: ages 22–35, various professions (e.g., software engineer, barista, chef), and income types (hourly, monthly, business-owner). For example, Fig. \ref{fig:Prompt}, David Kim has his own agent persona with his own demography and preferences, and so do other agents. The diversity underpins agent heterogeneity in behavioral triggers such as hunger, fatigue, or budget constraints, and leads to emergent group-level dynamics. Financial awareness, food availability, and proximity all contribute to different decision thresholds across agents.

\vspace{1em}

\subsubsection{Needs System and Internal Triggers} 
We simulate internal needs via a grocery–energy–finance triad. 
For example:

\begin{itemize}
  \item Grocery levels deplete after each meal and trigger shopping if agents drop below a threshold.
  \item Energy levels decrease with physical movement or work and can be replenished via rest or food.
  \item Money constraints affect purchasing decisions; agents can skip meals or choose cheaper options.
\end{itemize}

These needs feed directly into prompts and decision logic, ensuring agents are not only LLM-driven but also embodied with bounded rationality.
\subsection{Parallelism and Real-Time Multi-Agent Tracking}
\subsubsection{Parallelism}
This simulation integrates parallel execution with thread-safe memory and location management, enabling agents to behave simultaneously and independently within a shared environment. Rather than sequential turn-taking, the simulation operates as a sandbox where agents travel, make decisions, and interact in real time.

The system uses self-locks and location locks which ensure that:

\begin{itemize}
  \item Only one agent can access or modify its own memory state at a time (via a thread-safe self.\_lock mechanism)
  \item Only one agent can update a given location state at any moment (e.g., entering a shop or home).
\end{itemize}

This mechanism not only safeguards against concurrency errors, but also ensures that the majority of simulation time is spent on meaningful LLM-based reasoning and interaction, rather than bottlenecking on internal mechanics. As a result, the simulation supports higher agent counts and more dynamic social activity while maintaining coherence and speed.

Parallel execution with a thread-safe memory and location tracker allows agents to act simultaneously within a shared environment, enabling realistic real-time interactions. This design prevents conflicts when agents update their own memories or shared locations, supporting higher agent counts and more dynamic social activity without prescripting.

\vspace{1em}

\subsubsection{Shared Location Tracker} Real-time agent positioning
 \textbf{Shared Location Tracker}, is a critical enabler of this parallel system, which functions as a GPS for all agents. It maintains up-to-date location records across the simulation and allows agents to determine who is present at any given location. This is essential for triggering agents' spontaneous conversations, shared meals, or accidental encounters.

When agents move, the location tracker updates their positions using thread-safe lock, ensuring that no overlap or data collision occurs even under simultaneous execution. This system enables agents to interact organically, as they would in a real town, without requiring predefined scripts or manual scheduling.

\vspace{1em}
\begin{quote}
\footnotesize
\textbf{Tracker Interaction Flow:} \\
\texttt{Travel Action $\rightarrow$ Acquire Travel Lock $\rightarrow$ Update Shared Location $\rightarrow$ Log Memory $\rightarrow$ Release Lock}
\end{quote}
\vspace{1em}

By combining this real-time tracking with parallel LLM planning, the simulation supports behaviors such as running into acquaintances at a café, avoiding crowded spots, even choosing a quiet place to rest, all grounded on shared location tracker.

\subsection{Agent Memory and Conversational Dynamics}
To enable agents to exhibit believable social behavior and continuity in decision-making, we implement a structured memory system integrated with an LLM-powered conversational engine as Fig. \ref{fig:Prompt}. Each agent maintains a personalized memory stream that stores recent interactions, personal experiences, and contextual updates over time. These memories are selectively retrieved to guide future dialogue and planning. The memory management system consists of three key components:

\subsubsection{Memory Ingestion and Categorization}

Each significant observation-such as an activity, location change, emotional reflection, or conversation—is logged as a memory entry. These entries are tagged with metadata, including timestamps, memory type (EVENT, REFLECTION, CONVERSATION, PURCHASE), and the source agent’s name, enabling efficient filtering and recall during interactions.

\subsubsection{Memory Retrieval during Prompting}

When an agent engages in a new conversation or decision, the memory manager applies a time-decay filter and relationship proximity logic to retrieve the most relevant memory slices. This curated context is then embedded into the LLM prompt, allowing the model to produce consistent and socially grounded responses.

\subsubsection{LLM Conversation Integration} 

Through a centrally managed prompt system, agents receive structured prompts containing their current state, emotional history, and social connections. LLM responds in JSON format, including both dialogue and structured intent (e.g., travel, purchase, or event coordination). This enables agents to form evolving relationships, maintain routine behaviors, and even remember previously discussed plans. For example, an agent who had earlier talked about going for coffee may recall this memory and naturally act on it:
\vspace{1em}
\begin{quote}
\footnotesize
\ttfamily
\raggedright
Lisa mentioned going for coffee today.\\ 
Lisa asks: "Should we still go?"\\
David replies: "Yeah, I remember! Let's meet at the Local Café at 9 AM."\\
\end{quote}
\vspace{1em}

Rather than being hard-coded, these interactions emerge from the interplay of the memory stream and LLM response system, which together support realistic behaviors including routine formation, spontaneous coordination, and lapses in memory. Fig.\ref{fig:Flow Chart} above also illustrates how we ensure coherence between conversation and action, commitments made during dialogue are immediately stored commitments into memories and parsed into future plans. This allows both agents to update their internal schedules, ensuring synchronized behavior across future simulation ticks. Importantly, the system supports realistic deviations: an agent may forget, reschedule, or even change its mind based on updated memories or priorities.

\vspace{1em}
\begin{quote}
\footnotesize
\textbf{Interaction Flow:}
\vspace{1em}

\texttt{LLM Prompt$\rightarrow$Conversation Output$\rightarrow$Memory Update$\rightarrow$Execution Plan$\leftrightarrow$Retrieve Memory Stored}
\end{quote}
\vspace{1em}

Besides, the conversation frequency is not by random or hardcoded too, it's naturally decided by agents, conversation could be rejected due to reasons of low-energy, busy, or rushing to destination, etc. Here is one example through simulation process output:
\vspace{1em}
\begin{quote}
\footnotesize
\texttt{[DEBUG]Kevin Chen successfully completed activity 'work' at Fried Chicken Shop}
\texttt{[DEBUG] Social check skipped for Kevin Chen - busy with working}
\end{quote}
\vspace{1em}
This closed-loop architecture ensures that generative cognition is tightly integrated with agent behavior, forming a stable foundation for simulating believable, socially responsive agents.
\begin{figure*}[t]
    \centering
    \includegraphics[width=0.85\linewidth]{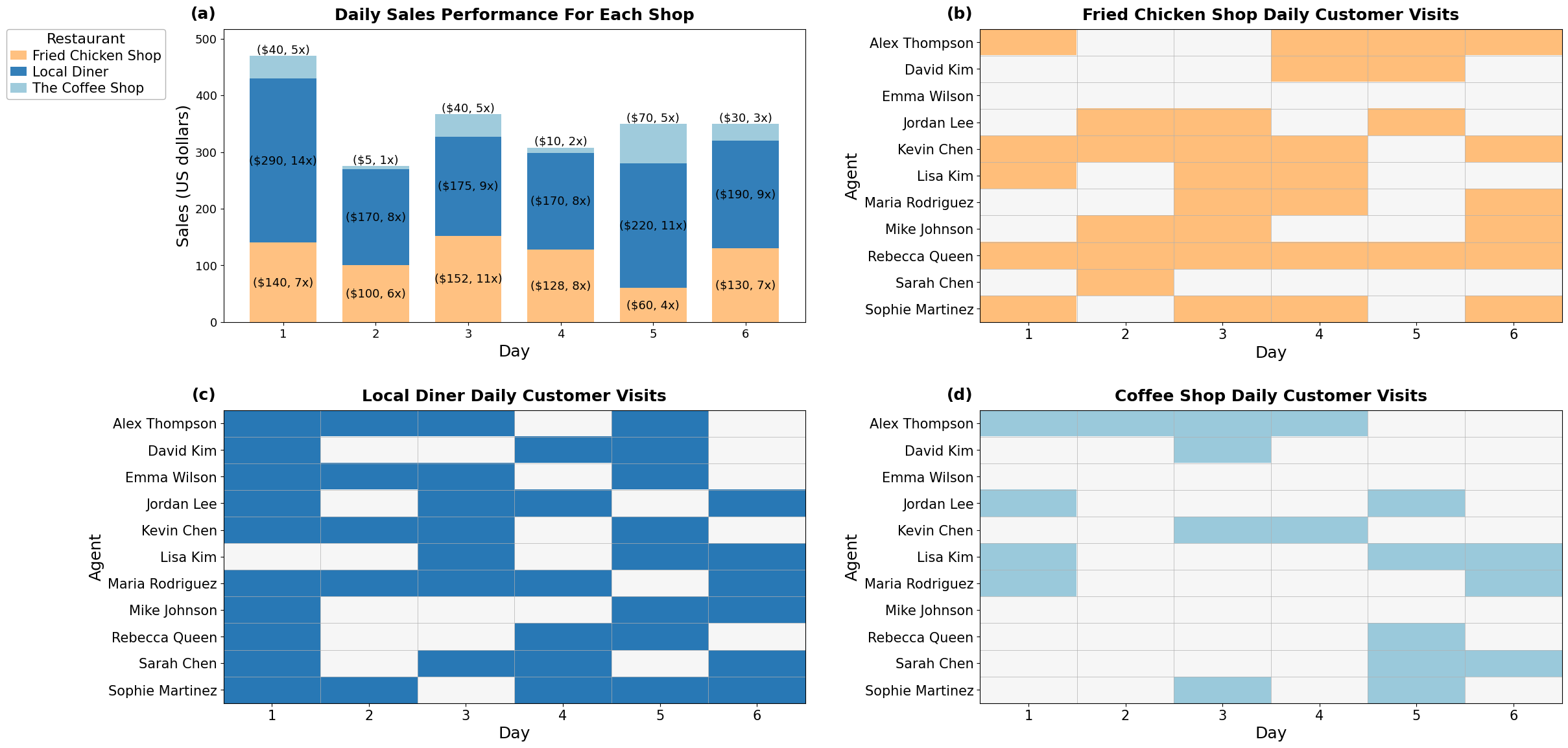}
    \caption{Daily Sales Performance Analysis and Customer Loyalty Through Visit Frequency}
    \label{fig:placeholder}
\end{figure*}
\subsection{Purchase Logic and Behavioral Constraints}
Agent purchases are not arbitrary; they emerge from a blend of internal needs, environmental context, financial capacity, and past experiences. These decisions are shaped through a structured reasoning pipeline and reinforced by memory, location logic, and budget constraints.

\subsubsection{From Need to Action}
Fig. \ref{fig:Prompt} tells the various prompt type management for different scenario in simulation. Low agent food or grocery supplies (e.g., post-breakfast-skip) activate the LLM Prompting System in response to physiological needs. The agent generates a natural-language reasoning and a structured plan to address the need. Besides, agent is required to plan and execute every meal (breakfast, lunch, dinner are required and snack optional), otherwise, they will miss energy due to the energy system's natural decay (imitating the human being system, costs energy for maintaining body function) for example:

\vspace{1em}
\begin{quote}
\footnotesize
\ttfamily
\raggedright
"decision": "Buy a meal from the Fried Chicken Shop",\\
\vspace{0.2em}
"reasoning": "I skipped breakfast and I’m nearby.",\\
\vspace{0.2em}
"action\_type": "purchase",\\
\vspace{0.2em}
"target\_location": "Fried Chicken Shop"\\
\end{quote}
\vspace{1em}
This decision only becomes executable if spatial and financial conditions are met. If not already at the shop, the agent schedules a travel action first. If lacking funds, the plan may be skipped, adjusted (e.g., choosing a cheaper option), or deferred entirely.

\subsubsection{Budget Awareness and Behavior Adjustment} 
Each agent operates under either monthly or hourly income for working hours, and maintains a running balance. These financial dynamics influence how often and where agents choose to eat. Some may favor cheaper diners, while others splurge at premium locations when paychecks arrive. If a proposed action exceeds their available budget, the agent may drop or revise the plan, reinforcing realism through bounded rationality.

Each agent determines the price of a potential purchase based on the item’s base cost and any applicable discount offered by the vendor. The final price \( \text{FinalPrice}_{i} \) for item \( i \) is calculated as follows:

\[
\text{F}_i = P_{\text{base},i} \times (1 - D_i)
\]
\noindent\textbf{Where:}
\begin{itemize}
  \item \( \text{F}_{i} \): Final price agent paid for item \( i \)
  \item \( P_{\text{base}, i} \): Base price of item \( i \), pre-defined in the shop's configuration menu
  \item \( D_{i} \): Discount rate applicable to item \( i \), defined in the configuration as a decimal (e.g., 20\% = 0.20)
\end{itemize}

This pricing logic is embedded into the agent’s purchase planning process. When agents evaluate food or product options during daily decision-making, they apply this formula to determine whether the item is affordable given their current budget. If the computed \( \text{FinalPrice}_{i} \) exceeds the agent's available funds, the agent may delay, cancel, or switch to a cheaper option. 

The mechanism ensures that purchase behaviors reflect realistic consumer decision patterns, including sensitivity to price reductions, bounded rationality, and budget constraints.

\vspace{1em}
\subsubsection{Memory-Informed Decisions}

Purchase behavior is also socially and personally shaped. For example, an agent might recall conversation memories:
\vspace{1em}
\begin{quote}
\footnotesize
\ttfamily
"Lisa mentioned about the Fried Chicken Shop last night. I’m curious to try it."\
\end{quote}

This memory is retrieved during planning and influences where the agent chooses to eat. Similarly, commitments made during conversations such as agreeing to lunch with a friend, are stored and translated into planned purchase actions when appropriate. In this way, memory fuels habit formation and social coordination.
\vspace{1em}
\subsubsection{Post-Purchase Feedback and Logging}
Once executed, purchases impact internal states (e.g., restoring grocery or energy levels), and are logged with details like time, cost, and location. As Fig.~\ref{fig:flow}, these logs feed into aggregated sales metrics and support downstream analysis of discount strategy effects. Flow Example:
\vspace{1em}
\begin{quote}
\footnotesize

\texttt{Hunger Foresee or Detected} \\
$\rightarrow$ \texttt{Plan to eat at Fried Chicken Shop} \\
$\rightarrow$ \texttt{Recall Lisa prefers the Local Caf\'e} \\
$\rightarrow$ \texttt{Change plan} \\
$\rightarrow$ \texttt{Check wallet} \\
$\rightarrow$ \texttt{Enough money} \\
$\rightarrow$ \texttt{Travel} \\
$\rightarrow$ \texttt{Purchase meal} \\
$\rightarrow$ \texttt{Update state (money, energy)}
\end{quote}
\vspace{1em}
This flow-grounded in needs, shaped by memory, and checked against environment, ensuring purchasing behaviors are not only diverse and emergent but also structurally coherent within the simulation system.
\section{Results and Analysis}

\subsection{Price Discount Impact on Sales Performance}

The 20\% midweek discount at Fried Chicken Shop produced measurable changes in market dynamics, as shown in Fig. \ref{fig:placeholder}. The shop's revenue increased 51\% from Day 2 (\$100.6) to Day 3 (\$152.11) despite the price reduction, indicating strong consumer response to the promotion. At the same time, Local Diner has decreased revenue for 7\%. This revenue enhancement persisted through Day 4 before declining, suggesting a promotional effect lasting approximately 2-3 days.

The discount triggered market share redistribution, with Fried Chicken Shop's share increasing from 30\% (Day 1) to 41\% (Day 3), while Local Diner's share decreased from 62\% to 48\%. Coffee Shop maintained stable performance throughout (\$30-40 daily), suggesting it serves a distinct customer segment less sensitive to lunch/dinner promotions. In the post-discount period (Days 5-6), Fried Chicken Shop's market share varied between 17\% and 37\%, showing volatility but maintaining presence above some pre-promotion levels.

The delayed peak on Day 3 rather than immediate response reveals gradual information diffusion through the agent network. Total daily market size fluctuated between \$276 - 471 without systematic expansion during the promotion period, indicating the discount primarily drove substitution between restaurants rather than increasing overall food consumption. This substitution effect, where customers shifted between providers without increasing total spending, demonstrates realistic consumer behavior in response to localized price promotions.

\subsection{Literature Validation}
These simulated effects align with long-standing marketing and economics findings. Classic studies show that temporary sales concentrate purchases within the promotional window and shift market share without expanding total demand \cite{salop1982sales, shoemaker1979promotions, hendel2006sales}. Our results exhibit the same substitution-driven pattern, supporting the validity of the generative simulation.

\subsection{Consumer Loyalty and Habit Formation}

The Shop Choice Matrix in Fig. \ref{fig:placeholder} reveals distinct patterns of consumer loyalty and repeat visitation across the simulation period. Analysis of individual agent trajectories demonstrates both promotion-driven and organic loyalty formation, with notable differences between establishments.

Fried Chicken Shop exhibited strong loyalty patterns during the discount period by certain old customers (Kevin, Rebecca, Sophie) consecutive visits. while Alex, Jordan and Sophie demonstrated repeat patronage in the latter period. This temporal clustering coincides with the promotional period, suggesting the discount not only attracted new customers who never visited before, but also prompted previous customers to visit more frequent. The pattern indicates habit formation, as repeat visits occurred even as the promotional novelty diminished.

In contrast, Local Diner demonstrated broader appeal across the agent population without promotional incentives. Multiple agents including Emma Wilson, David Kim, and Kevin Chen showed multi-day patronage patterns. This distributed loyalty, occurring without price manipulation, suggests Local Diner maintained consistent appeal as a default dining choice for various agent.

The emergence of consensus behavior without fixed programming validates the simulation's fidelity. Multiple agents independently converged on Local Diner as a frequent choice, mirroring real-world phenomena where certain establishments achieve preferred status. The matrix reveals heterogeneous patterns: some agents like Maria Rodriguez explored options before routine formation while others maintained diverse choices. This loyalty variety formation encompasses exploratory, habitual, and variety-seeking behaviors, demonstrating that LLM-powered agents replicate consumer decision strategies without deterministic rules.

\begin{figure}[h]
    \centering
    \includegraphics[width=0.75\columnwidth]{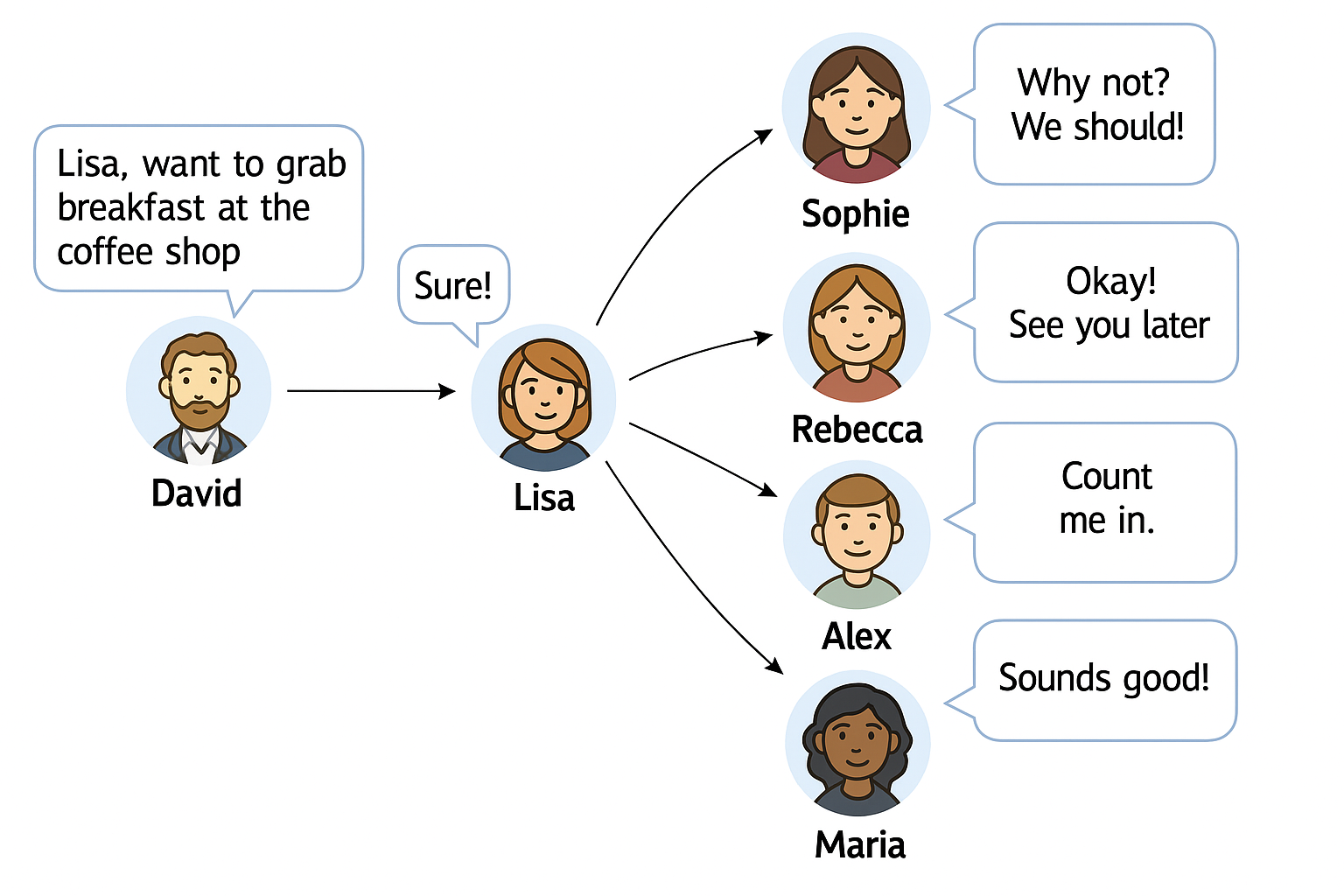}
    \caption{Agent Lisa makes a breakfast plan with David about visiting the Coffee Shop. Lisa makes few more conversations with invitations that could potentially become a large group gathering.}
    \label{fig:multi-agent-coordination}
\end{figure}

\subsection{Emergent Social Dynamics and Multi-Agent Coordination}

Beyond individual purchase decisions, the simulation revealed unexpected social coordination patterns that emerged without explicit programming. Fig. \ref{fig:multi-agent-coordination} illustrates a case where Agent David (Lisa's husband) initiated a breakfast plan to Lisa at the Coffee Shop (closest relationship invitation), then Lisa extended similar invitations to multiple other agents (Sophie, Rebecca, Alex, and Maria) through separate conversations, respectively are her friends and acquaintances talked before. While Sophie agreed ("Sure!"), Rebecca suggested meeting later ("Okay! See you later"), Alex showed enthusiasm ("Count me in"), and Maria expressed interest ("Sounds good!"), demonstrating varied but realistic social responses.

This multi-invitation pattern emerged through the interaction of memory and conversation systems, which allowed agents to recall prior commitments and maintain consistency across dialogues. Meanwhile, simulation mirrors real-world social hubs and demonstrates how word-of-mouth diffusion can arise organically through peer-to-peer exchanges, highlighting the framework’s potential for modeling viral marketing and consumer network effects such as experiencing marketing strategy or trendy phenomenon on social media.

\section{Known Limitation}
\subsection{LLM Sensitivity and Prompt Robustness}

A notable challenge in LLM-driven simulations is the model's tendency to hallucinate responses outside the configured environment, even when provided with structured context. For instance, our daily planning prompt is approximately 7,098 characters, 2225 characters for conversation in length, and includes detailed rules about the agent's current state(location, energy, money, etc), previous memory load back, location options, work routines, and agent constraints. Despite this, the LLM may still propose actions involving nonexistent locations:
\vspace{1em}
\begin{quote}
\footnotesize
\ttfamily
\raggedright
"time": 19,\\
"action": "eat",\\
"target": "new bistro near Oak View Condos",\\
"description": "Dinner plans with Sophie"\\
\end{quote}
\vspace{1em}

This output is problematic because "new bistro" doesn't exist in town map. To reduce such inconsistencies, we enforce prompt constraints and clarify rules directly within the instruction (e.g., “Only choose from the following known locations...”), but the risk of generative flexibility remains.

Furthermore, such behavior varies across LLMs. While DeepSeek-V3 demonstrated relatively stable compliance, other models like llama2 occasionally diverged under identical prompt structures—suggesting that simulation reproducibility is not only prompt-dependent but also model-specific. This introduces an additional layer of sensitivity that must be considered in longitudinal experiments or comparative studies.

We consider the need for a formal grounding mechanism, or constrained decoding strategy, a promising direction for future work. Techniques such as location entity checking, model ensemble verification, or hybrid symbolic + generative logic could further enhance planning fidelity.

\subsection{Execution Sensitivity and Energy Recovery Handling}
Despite structured prompting and memory integration, we observed execution sensitivity issues when similar words or misunderstanding appear:
\vspace{1em}
\subsubsection{Ambiguity in Intent and Resource Interpretation} 
Agents may confuse \textbf{qualitatively different activities} (e.g., "have breakfast" vs. "grab a coffee") or invent unintended plans:
\begin{itemize}
  \item An agent may drink only coffee, assuming it satisfies a meal, which fails to restore enough energy.
  \item Agents might plan to "go buy a smoothie" despite no such item being listed in the shop’s available menu. 
\end{itemize}
These cause execution errors during the simulation tick, where the environment cannot fulfill the agent’s plan.
\vspace{1em}
\subsubsection{Energy Decay and Undetected Starvation}
To maintain physiological realism, agents experience natural energy decay per hour, even when idle (e.g., simulating basal metabolic rate). However, if agents:
\begin{itemize}
  \item Miss multiple meals due to misaligned planning
  \item Fail to correctly purchase available food (due to invented or unrecognized targets in LLM daily plan response)
\end{itemize}
The agents will experience progressive energy depletion. When energy level reaches 0, agents are marked as non-functional, unable to travel or act—what we call a \textbf{“dead agent” state}. To prevent simulation collapse, we implement the following fallback logic:
\begin{itemize}
  \item If an agent's energy drops below a critical threshold (energy = 20 here), the system triggers an emergency re-plan using a stricter, food-focused prompt to prioritize recovery.
  \item If this fails and energy reaches 0, the agent is force-teleported to their residence, simulating collapse or rest.
  \item During the night, all agents undergo energy restoration during sleep, allowing them to recover and rejoin the simulation the next morning.
\end{itemize}
This design prevents long-term simulation truncation, especially when running large-scale, multi-agent environments, where such issues are otherwise hard to detect in real time.
\vspace{1em}
\subsection{Age-Specific Behavioral Fidelity}
Although our simulation framework supports agent profiles spanning different ages, occupations, and socioeconomic statuses, we observed fidelity gaps when modeling children and older adults. These limitations likely stem from LLM training data, which are dominated by internet users aged roughly 18–45. For example, a 7-year-old agent assigned childlike attributes still displayed adult behaviors (e.g., requesting coffee when tired) and lacked age-appropriate curiosity or emotional tone. Similarly, elderly agents showed little minor distinction from middle-aged adults, rarely referencing their health concerns, technological unfamiliarity, or traditional perspectives.
These limitations reduce the realism of simulations targeting edge demographics and may hinder the study of:
\begin{itemize}
\item Educational or entertainment products for children
\item Marketing and UX design for elderly consumers
\item Multi-generational family or care scenarios
\end{itemize}
Future work may benefit from persona-specific adapters or age-targeted finetuning to ensure behavioral authenticity beyond the median internet user.

\section{Conclusion}

Our study advances generative agent frameworks by simulating consumer decision-making through LLM-powered agents in a virtual town. The architecture integrates economic reasoning, spatial planning, and social memory capturing both expected phenomena such as discount-driven purchase surges and emergent behavioral dynamics. We demonstrate group coordination, habit formation, and conversational influence alongside unprogrammed patronage patterns where agents consistently favored popular establishments, exhibiting social clustering that reflects real consumer psychology.


These behaviors mirror human tendencies, demonstrating that LLMs generate socially aligned, high-fidelity behavior autonomously. The realism and combinability position LLM simulations as bridges between AI advancement and strategic experimentation, enabling generative AI to serve as both modeling tools and campaign testbeds. The framework extends beyond discount tests to multichannel and longitudinal scenarios, advancing e-business innovation.

\section*{Ethical Considerations and Bias Mitigation}
We carefully balances agent personas across gender, occupation, and income categories to reduce systematic privilege and create a diverse market environment. However, the underlying large language model is trained on internet-scale data that may embed demographic biases or cultural stereotypes, which can influence agent behavior and result interpretation in ways that are difficult to fully control through simulation design alone. To mitigate these risks, outputs were regularly reviewed for biased language and prompt constraints were adjusted when problematic patterns appeared. Nevertheless, the findings should be interpreted as illustrative of behavior within this constrained setting, not as evidence of real-world demographics. Future work could explore alternative or fine-tuned LLMs and expand the range of personas to capture more nuanced characteristics, further balancing social factors and moving the simulation closer to the complexity of real-world consumer environments.

\section*{Data Availability}

The source code and experimental data are available at: \url{https://github.com/carolchu1208/LLM-Based-Generative-Agents-Simulating-Consumer-Decisions}
\section*{Acknowledgment}
The authors thank Prof. Zonghao Yang from Stevens Institute of Technology for fruitful discussions and valuable feedback on this work.

\end{document}